\documentclass[11pt,letterpaper]{article}
\usepackage{naaclhlt2015/naaclhlt2015}
\usepackage{times}
\usepackage{latexsym}

\usepackage{url}

\usepackage[utf8x]{inputenc}
\usepackage[pdftex]{graphicx}
\usepackage[all]{xy}
\usepackage{amsmath,amssymb,multirow,color}
\usepackage{algorithmic}
\usepackage[algoruled,linesnumbered,noend,noline]{algorithm2e}
\usepackage{multirow}

\usepackage{dashrule}
\usepackage{enumitem}

\usepackage{rotating}
\usepackage{cleveref}

\title{Automatic Prediction of the Performance of Every Parser}

\author{Ergun Bi\c{c}ici \\
              ADAPT CNGL Centre for Global Intelligent Content \\
              School of Computing \\
              Dublin City University, Dublin, Ireland. \\
              {\tt ergun.bicici@computing.dcu.ie}
}

\date{}

\begin{document}

\maketitle

\begin{abstract}
We present a new parser performance prediction (PPP) model using machine translation performance prediction system 
(MTPPS), statistically independent of any language or parser, relying only on extrinsic and novel features based on 
textual, link structural, and bracketing tree structural information. 
This new system, MTPPS-PPP, can predict the performance of any parser in any language
and can be useful for estimating the grammatical difficulty when understanding
a given text, for setting expectations from parsing output, for parser selection for a specific domain,
and for parser combination systems.
We obtain SoA results in PPP of bracketing $F_1$ with better results over textual features and similar 
performance with previous results that use parser and linguistic label specific information.
Our results show the contribution of different types of features as well as rankings of individual
features in different experimental settings (cased vs. uncased), in different learning tasks (in-domain vs.
out-of-domain), with different training sets, with different learning algorithms, and with different dimensionality 
reduction techniques.
We achieve $0.0678$ MAE and $0.85$ RAE in setting +Link, which corresponds to about $7.4\%$ error when 
predicting the bracketing $F_1$ score for the Charniak and Johnson parser on the WSJ23 test set. 
MTPPS-PPP system can predict without parsing using only the text, without a supervised parser using only an 
unsupervised parser, without any parser or language dependent information, without using a reference parser output, and 
can be used to predict the performance of any parser in any language.
\end{abstract}

\section{MTPPS-PPP}

Parsing gives an insight into how a sentence is formed and how its structural components interact
when conveying the meaning.
Parser training and parsing can be computationally demanding and for supervised
parsers, training requires labeled tree structural data, which can be expensive to obtain,
scarce, or limited to a single domain. 
PPP is useful for estimating the grammatical difficulty when understanding a given 
text, for setting expectations from parsing output, and it can be used for parser selection for a specific domain and 
for parse selection from the output of multiple parsers on the same data. 
PPP before parser training and parsing can prevent spending resources towards obtaining training data and parser 
training. PPP after parsing can help us select better parsers for a given domain.

MTPPS as described in~\cite{BiciciMTPPMTJ,Bicici:RTM4QE:WMT13,Bicici:RTM4QE:WMT14} provide parser and language 
independent features, which can be used to predict the performance 
of any parser in any language, measuring the closeness of a given sentence to be parsed to the training set available, 
the difficulty of retrieving close sentences from the training set, and the difficulty of translating them to a known 
training instance. 
MTPPS-PPP features measure similarity of the grammatical structures, the vocabulary, and their distribution using a 
subset of the MTPPS features and some additional features:

\begin{description}[leftmargin=0.35cm,labelindent=0.0cm]
\item[Text \{121\}] use $n$-grams as the basic units of information over which similarity 
calculations are made with up to $3$-grams for textual features and $5$-grams for LM features.
Textual features allow us to predict without actually training the parser or parsing with it. 

\item[Link \{26\}] use link structures from unsupervised parser CCL~\cite{SeginerThesis}, which
has linear time complexity in the length of a sentence~\cite{SeginerThesis}. Derivation of the textual 
features for PPP take minutes. Link contain coverage \{9\}, perplexity \{15\}, and $\chi^2$ vector similarity \{2\} 
features.
\end{description}

\vspace*{-0.5cm}
\begin{figure}[th]
\centering
\includegraphics[width=0.8\linewidth]{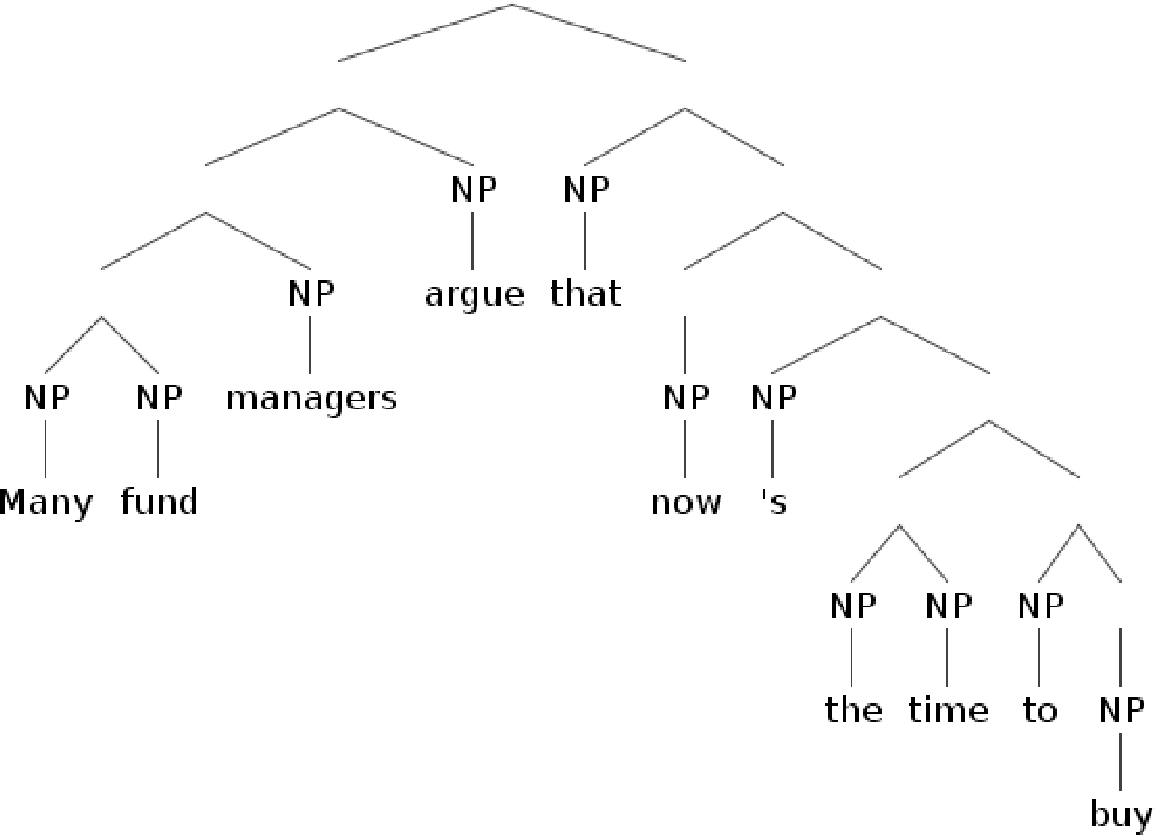}
{\scriptsize
\begin{tabular}{@{\hspace{0.0cm}}c@{\hspace{0.01cm}}c@{\hspace{0.01cm}}c@{\hspace{0.01cm}}c@{\hspace{0.01cm}}c@{\hspace{
0.0cm}}} 
\multicolumn{5}{c}{CCL} \\
numB & depthB & avg depthB & R/L & avg R/L \\ \hline24.0 & 9.0 & 0.375 & 2.1429 & 3.401 
\vspace*{-0.35cm}
\\
\parbox{0.15\linewidth}{\scriptsize\begin{displaymath} \xymatrix@=0.3cm{2 \ar@{-}[d] \ar@{-}[dr] &  \\1 & 1} 
\end{displaymath}} & \parbox{0.15\linewidth}{\scriptsize\begin{displaymath} \xymatrix@=0.3cm{1 \ar@{-}[d] \ar@{-}[dr] & 
\\1 & 13} \end{displaymath}} & \parbox{0.15\linewidth}{\scriptsize\begin{displaymath} \xymatrix@=0.3cm{1 \ar@{-}[d] 
\ar@{-}[dr] &  \\1 & 2} \end{displaymath}} & \parbox{0.15\linewidth}{\scriptsize\begin{displaymath} \xymatrix@=0.3cm{1 
\ar@{-}[d] \ar@{-}[dr] &  \\1 & 8} \end{displaymath}} & \parbox{0.15\linewidth}{\scriptsize\begin{displaymath} 
\xymatrix@=0.3cm{1 \ar@{-}[d] \ar@{-}[dr] &  \\2 & 10} \end{displaymath}} \vspace*{-0.7cm} \\
\parbox{0.15\linewidth}{\scriptsize\begin{displaymath} \xymatrix@=0.3cm{1 \ar@{-}[d] \ar@{-}[dr] &  \\3 & 1} 
\end{displaymath}} & \parbox{0.15\linewidth}{\scriptsize\begin{displaymath} \xymatrix@=0.3cm{1 \ar@{-}[d] \ar@{-}[dr] & 
\\3 & 4} \end{displaymath}} & \parbox{0.15\linewidth}{\scriptsize\begin{displaymath} \xymatrix@=0.3cm{1 \ar@{-}[d] 
\ar@{-}[dr] &  \\5 & 1} \end{displaymath}} & \parbox{0.15\linewidth}{\scriptsize\begin{displaymath} \xymatrix@=0.3cm{1 
\ar@{-}[d] \ar@{-}[dr] &  \\7 & 15} \end{displaymath}} & 
\vspace*{-0.35cm}
\end{tabular}
}
\caption{Tree features for a CCL parsing output (immediate non-terminals replaced with NP).}
\label{TreeFeatures}
\end{figure}

\begin{description}[leftmargin=0.35cm,labelindent=0.0cm]
\item[TreeF \{5\}] use parse trees to derive features based on their bracketing structure (Figure~\ref{TreeFeatures}). 
TreeF provide information about the structural properties of the parser outputs such as the link structural information
and geometrical properties of the output bracketing structure.
We derive $5$ statistics: number of brackets used (numB), depth 
(depthB), average depth (avg depthB), number of brackets on the right branches over the number of brackets on the left 
(R/L),\footnote{For nodes with an uneven number of children, the nodes in the odd child contribute to the right 
branches.} average right to left branching over 
all internal tree nodes (avg R/L). The ratio of the number of right to left branches shows the degree to which the 
sentence is right branching or not. Additionally, we capture the different types of branching present in a given parse 
tree identified by the number of nodes in each of its children. 
These represent the decomposition of the branching structure of the parse tree. Decomposition allows 
reconstruction of the branching structure and similarity measurements at a finer level. 
\end{description}

\begin{description}[leftmargin=0.35cm,labelindent=0.0cm]
\item[C$F_1$ \{1\}] use comparative $F_1$ score, which is the average relative bracketing $F_1$ of each parser output: 
$CF_1(p) = \frac{1}{|P|-1} \sum_{p_i \in P, p_i \neq p} F_1(p, p_i)$.\footnote{Ravi et al.~\cite{SoricutAPP2008} use an 
additional supervised parser output to calculate bracketing $F_1$ score as an additional feature. However, using a 
top performing parser output such as that from CJ, which can achieve $0.91$ bracketing $F_1$ 
(\Cref{BaselineParserPerformance}) as the reference, can be close to looking at the reference itself.}
\end{description}

MTPPS-PPP requires minutes to derive the features and 
predict and solves an important language processing problem with a large application area in significantly
less computational effort.\footnote{Compared to the time that would be needed for gathering training data
and training parsers.}

Ravi et al.~\cite{SoricutAPP2008} focus on predicting the performance of a single parser, Charniak and Johnson 
(CJ)~\cite{charniak-johnson:2005:ACL}, using text based and parser based features, and reference parser output (Bikel 
parser~\cite{Bikel2002}). 
The text features are the length of the sentences, a bag of words for discriminating easy and hard to parse sentences 
selected based on information gain, the number of unknown words, and the LM perplexity. 
Parser based features include the syntactic category of the root of the parser output tree, the number of nodes tagged 
as punctuation, and the number of nodes that are tagged as a particular label (among $72$ labels). Parser based 
features are dependent on the labels used in the parsing output as well as the labels used in the training set and 
therefore not applicable for parser independent evaluation and prediction. Reference parser output is used to score the 
parser output and obtain bracketing $F_1$ scores.

In contrast to related work, we develop a model for parser independent evaluation and prediction of performance 
including the prediction of unsupervised parser performance. 
MTPPS-PPP prevents dependence on any task, domain, language, or parser dependent information and works in a 
realistic scenario where parser outputs may be provided by different parsers possibly trained on different training 
data. 
We achieve better results using only textual features and obtain similar results to the previously published
results relying on only extrinsic features and without any parser or label dependent information or without
using features based on a top performing reference parser output.
Using the output of a reference parser can defeat the purpose of predicting the performance with small computational 
cost and a SoA reference parser that can reach $\geq 0.9$ $F_1$ score may be close to using the
reference itself. MTPPS-PPP can predict without parsing using only the text, without training a supervised parser 
using only the text for training an unsupervised parser, without any parser or language dependent information, and 
without using a reference parser output. 

\section{Statistical Lower Bound on Error}
\label{statisticalLowerbound}

We evaluate the prediction performance with correlation ($r$), root mean squared error (RMSE), mean absolute error 
(MAE), or relative absolute error (RAE). 
We obtain expected lower bound on the prediction performance, number of instances needed for prediction given a 
RAE level and for building SoA predictors for a given prediction task
From a statistical perspective, we can predict the number of training instances we need for learning with a goal of
increasing the signal to noise ratio $\mbox{SNR} = \frac{\mu}{\sigma}$ or the ratio of the mean to the
standard deviation.
Let $\textbf{y} = (y_1, \ldots, y_n)^T$ represent the target sampled from a distribution with mean $\mu$ and standard
deviation $\sigma$, then the variance of $\sum_{i=1}^n y_i$ is $n \sigma^2$ and of the sample mean,
$\bar{\textbf{y}}$, is $\frac{\sigma^2}{n}$ with the standard deviation becoming $\frac{\sigma}{\sqrt{n}}$.
Thus, by increasing the number of instances, we can decrease the noise in the data and increase $\mbox{SNR}$. 
We are interested in finding a confidence interval, $[\bar{\textbf{y}} - t \frac{s}{\sqrt{n}}, \bar{\textbf{y}} + t
\frac{s}{\sqrt{n}}]$, where the value for $t$ is found by the Student's $t$-distribution for
$n-1$ degrees of freedom with confidence level $\alpha$. True score lies in the interval with probability $1-\alpha$ or 
for 
$s$ representing the sample standard deviation of the scores:\footnote{This forms the
basis for many statistical significance tests in machine translation~\cite{BiciciThesis}.}
\begin{equation}
P(\bar{\textbf{y}} - t \frac{s}{\sqrt{n}} \leq \mu \leq \bar{\textbf{y}} + t \frac{s}{\sqrt{n}}) = 1 - \alpha.
\end{equation}
The absolute distance to the true mean or the width of the interval, $d$, is empirically equal to MAE and the 
relationship between RAE and MAE is as follows:
\begin{eqnarray}
d = \frac{t s}{\sqrt{n}} \Rightarrow n = \frac{t^2 s^2}{d^2} \label{NoiseLevel} \\
\mbox{RAE} = \frac{n \mbox{MAE}}{\sum_{i=1}^n |\bar{\textbf{y}} - y_i|} \label{RAEEquation}
\end{eqnarray}
Using Equation~\ref{RAEEquation}, we can derive MAE for a given RAE as an estimate of $d$, $\hat{d}$.
We confidently estimate (with $\alpha=0.05$, $p=0.95$) $\hat{d}$ and the corresponding $\hat{n}$ to reach the 
required noise level for the prediction tasks given a possible RAE level using Equation~\ref{NoiseLevel}.

Statistical lower bound on PPP error is presented in~\Cref{TrainingSizesPPP}, which lists how many training 
instances to use for PPP.
\Cref{TrainingSizesPPP} presents the $d$ possible according to the bracketing $F_1$ score distribution and the training 
set sizes required for reaching a specified noise level based on the RAE. 
$\hat{n}$ increase linearly with RAE and we observe that every $1\%$ decrease in 
RAE correspond to tenfold increase in $\hat{n}$ corresponding to a slope around 10.

\begin{table}[t]
\centering
{
\begin{tabular}{@{\hspace{0.0cm}}c|c@{\hspace{0.2cm}}c|c@{\hspace{0.2cm}}c@{\hspace{0.0cm}}}
& \multicolumn{2}{c|}{WSJ24} &
\multicolumn{2}{c}{WSJ02-21} \\
\hline
n & \multicolumn{2}{c|}{1346} & \multicolumn{2}{c}{6960} \\
$\mu$ & \multicolumn{2}{c|}{0.7095} & \multicolumn{2}{c}{0.7145} \\
s & \multicolumn{2}{c|}{0.1636} & \multicolumn{2}{c}{0.1633} \\
d & \multicolumn{2}{c|}{0.0087} & \multicolumn{2}{c}{0.0038} \\ 
\hline
RAE & $\hat{d}$ & $\hat{n}$ & $\hat{d}$ & $\hat{n}$ \\
1\% & 0.0013 & 57335 & 0.0013 & 58164 \\
5\% & 0.0067 & 2296 & 0.0066 & 2329 \\
10\% & 0.0134 & 576 & 0.0133 & 584 \\
20\% & 0.0268 & 146 & 0.0265 & 148 \\
30\% & 0.0402 & 66 & 0.0398 & 67 \\
40\% & 0.0536 & 38 & 0.0531 & 39 \\
50\% & 0.0670 & 25 & 0.0664 & 26 \\
75\% & 0.1004 & 13 & 0.0995 & 13 \\
80\% & 0.1071 & 12 & 0.1062 & 12 \\
85\% & 0.1138 & 11 & 0.1128 & 11 \\
\hline
\end{tabular}
}\caption{Estimated $d$ and $\hat{d}$ and $\hat{n}$ required for the noise levels based
on RAE when predicting bracketing $F_1$.}
\label{TrainingSizesPPP}
\end{table}

Top results in quality estimation task for machine translation achieve RAE of 
$0.85$~\cite{Bicici:RTM4QE:WMT14} when predicting HTER, which is an 
evaluation metric with range $[0,1]$. 
We show that we can obtain RAE of $0.85$ in \Cref{WSJ23TestResults} when predicting bracketing $F_1$, where 
bracketing $F_1$ has also the range $[0,1]$. 
With only 11 labeled instances for PPP, we can reach SoA prediction performance. 

\begin{figure}[t]
\centering
\includegraphics[width=1.0\linewidth]{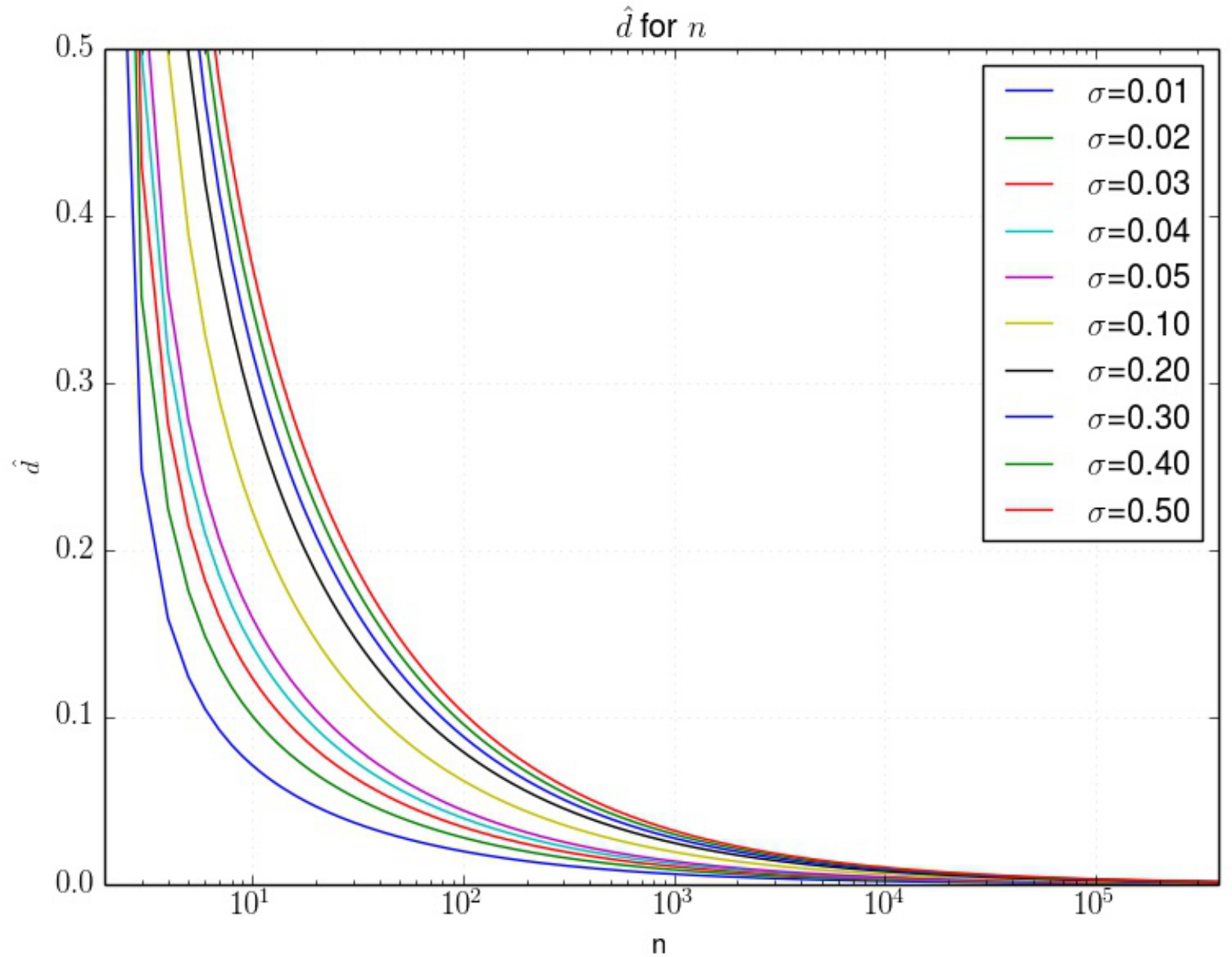}
\caption{Prediction error decrease by: (i) increasing $n$; (ii) decreasing $s$.}
\label{Figure_lowerdeviation_lowererror}
\end{figure}

\Cref{Figure_lowerdeviation_lowererror} samples from normal $n$-gram $F_1$~\cite{BiciciThesis} distributions with 
$\mu=0.2316$ from MTPPDAT\footnote{\texttt{https://github.com/bicici/MTPPDAT}} for different $\sigma$ and shows that 
prediction error decrease by: (i) increasing $n$; (ii) decreasing $s$.
MTPPDAT (MTPP dataset) contains document and sentence translation experiments collected from
4 different settings: tuning (TUNE), no tuning (NTUNE), multiple perspective learning
(MPL), and adaptation (ADAP)~\cite{BiciciPBML2014}. 

\section{Experiments}
\label{Experiments}

We present an extensive study of parser performance prediction using different types of features
(text, link, tree, C$F_1$), in different experimental settings (cased vs. uncased), and in different
learning tasks
(in-domain vs. out-of-domain).
We use the Wall Street Journal (WSJ) and Brown corpora distributed with Penn Treebank version
3~\cite{PennTB1993,LDCPennTB}.
WSJ02-21 refer to WSJ sections 2 to and including 21, WSJ24 refer to WSJ section 24, WSJ23 refer to
WSJ section 23, and
WSJ0,1,22,24 refer to WSJ sections 00, 01, 22, and 24 combined. BTest refers to the test set formed
by selecting every 10th sentence from the Brown corpus following~\cite{SoricutAPP2008}.
The number of sentences for each test set is given in \Cref{BaselineParserPerformance}. WSJ02-21
contains 39832 sentences in total and WSJ0,1,22,24 contains 6960 sentences.
We obtain the raw format for the Penn Treebank starting from the parse annotated
sentences.\footnote{On some cases,
sentences are split after white space and we don't know which are the correct ones to split after.}

\subsection{Parsers}

\textbf{CCL:} CCL~\cite{SeginerThesis} is an unsupervised parsing algorithm, which
allows equivalent classes with reciprocal links between words. \\
\textbf{PCFG:} Plain PCFG (probabilistic context free grammar) parser uses the Stanford
supervised parser~\cite{Klein2003}.\footnote{v2.0.3.} The plain PCFG model is unlexicalized; it has
context-free rules conditioned on only the parent nodes; it does not have language dependent heuristics for unknown
word processing; and it selects the left-most category as the head of the right hand side of a rule. \\
\textbf{CJ:} Charniak and Johnson~\cite{charniak-johnson:2005:ACL} develop a
SoA parser achieving the highest performance by reranking $50$ best parses with a
maximum entropy reranker.


\begin{table}[t]
{
\hspace*{1.2cm}
\begin{center}
\begin{tabular}{@{\hspace{0.0cm}}l|c@{\hspace{0.1cm}}c@{\hspace{0.1cm}}c@{\hspace{
0.1cm}}c@{\hspace{0.1cm}}c@{\hspace{0.0cm}}}
\hline
Corpus & numB & depthB & avg depthB & R/L & avg R/L \\
\hline
WSJ02-21 & 46.4 & 11.1 & 0.2678 & 6.46 & 6.68 \\
\hline
WSJ23 & 45.6 & 11.0 & 0.2728 & 6.36 & 6.66 \\
${S_{P_{\mbox{\tiny CCL}}}}^\prime$ & 38.6 & 9.3 & 0.2829 & 6.14 & 6.14 \\
${S_{P_{\mbox{\tiny PCFG}}}}^\prime$ & 41.6 & 10.0 & 0.2735 & 6.11 & 5.72 \\
${S_{P_{\mbox{\tiny CJ}}}}^\prime$ & 42.6 & 11.0 & 0.2887 & 5.96 & 6.27 \\
\hline
BTest & 38.1 & 9.6 & 0.3060 & 6.09 & 5.50 \\
${S_{P_{\mbox{\tiny CCL}}}}^\prime$ & 31.8 & 8.8 & 0.3551 & 6.77 & 6.04 \\
${S_{P_{\mbox{\tiny PCFG}}}}^\prime$ & 35.1 & 9.1 & 0.3165 & 7.05 & 5.25 \\
${S_{P_{\mbox{\tiny CJ}}}}^\prime$ & 35.6 & 9.7 & 0.3248 & 6.63 & 5.50 \\
\hline
\end{tabular}
\end{center}
}\caption{Tree structure statistics.}
\label{TreeFStats}
\end{table}

Parse tree branching statistics for WSJ2-21, WSJ23, and BTest together with the parser outputs obtained with different
parsers are given \Cref{TreeFStats}.
CCL output parse trees tend to have fewer branches and less depth. However, CCL outputs trees with
closer R/L and avg R/L to the test set than PCFG. 
CJ outputs trees with closest numB and depthB to the test sets. PCFG achieves the closest avg depthB.
The most frequent tree structure features may not be the best set of features to retain. Therefore,
we experiment with
feature selection (\Cref{FeatureSelectionResults}) over this set and in the final model, we use the
$5$ tree
statistical features together with the bag of tree structural features selected from the $70$ most
frequent found in
the training corpora.
\Cref{TreeFStats} shows that English is a predominantly right branching language.

\begin{table}[t]
{\small
\hspace*{1.2cm}
\begin{center}
\begin{tabular}{@{\hspace{0.0cm}}l@{\hspace{0.2cm}}c|c@{\hspace{0.2cm}}c@{\hspace{0.2cm}}c|c@{\hspace{ 
0.2cm}}c@{\hspace{0.0cm}}}
 & & \multicolumn{3}{c|}{Cased} & \multicolumn{2}{c}{Lowercased}
\\
Test & \# sents & CCL & PCFG & CJ & PCFG & CJ \\
\hline
WSJ24 & 1346 & 0.5489 & 0.6915 & 0.9057 & 0.6887 & 0.9008 \\
WSJ23 & 2416 & 0.5501 & 0.6933 & 0.9141 & 0.6901 & 0.9093 \\
BTest & 2425 & 0.5646 & 0.6773 & 0.8561 & 0.6748 & 0.8525 \\
\hline
\end{tabular}
\end{center}
}
\caption{Baseline parser performances trained on WSJ02-21 in terms of bracketing $F_1$ over all
sentences.}
\label{BaselineParserPerformance}
\end{table}

\begin{table*}[t]
{\small
\begin{center}
\begin{tabular}{@{\hspace{0.0cm}}l@{\hspace{0.2cm}}l|c@{\hspace{0.2cm}}@{\hspace{0.2cm
}}c@{\hspace{0.2cm}}r@{\hspace{0.2cm}}l@{\hspace{0.2cm}}l@{\hspace{0.2cm}}l@{\hspace{0.2cm}}l|c@{
\hspace{0.2cm}}c@{\hspace{0.2cm}}r@{\hspace{0.2cm}}l@{\hspace{0.2cm}}l@{\hspace{0.2cm}}l@{\hspace{0.2cm}}l@{\hspace{
0.0cm}}}
& & \multicolumn{7}{c|}{Train: WSJ24} & \multicolumn{7}{c}{Train: WSJ0-1-22-24} \\
Setting & Parser & \#dimI & Model & \#dim & $r$ &
RMSE & MAE & RAE & \#dimI & Model & \#dim & $r$ & RMSE & MAE & RAE \\
\hline
 \multirow{3}{*}{ Text }  & CCL & 132 & SVR & 132 & 0.48 & 0.1345 & 0.107 & 0.87 & 132 & SVR & 132 & 0.49
& 0.1333 & 0.1054 &
0.85 \\
  & PCFG & 132 & SVR & 132 & 0.32 & 0.1619 & 0.1239 & 0.93 & 132 & SVR & 132 & 0.36 & 0.1614 & 0.1218 & 0.91
\\
  & CJ & 132 & SVR & 132 & 0.27 & 0.1032 & 0.0702 & 0.88 & 132 & SVR & 132 & 0.27 & 0.1069 & 0.0681 & 0.86
\\
  \hline
  \multirow{3}{*}{ +Link }  & CCL & 158 & SVR & 158 & 0.48 & 0.1341 & 0.1067 & 0.87 & 158 & SVR-PLS & 44
& 0.5 & 0.133 & 0.1044
& 0.85 \\
  & PCFG & 158 & SVR & 158 & 0.3 & 0.1632 & 0.1245 & 0.93 & 158 & SVR & 158 & 0.36 & 0.161 & 0.1216 & 0.91
\\
  & CJ & 158 & SVR & 158 & 0.27 & 0.1031 & 0.0704 & 0.89 & 158 & SVR-FS & 4 & 0.23 & 0.1085 & 0.0678 & 0.85
\\
  \hline
  \multirow{3}{*}{ +TreeF }  & CCL & 172 & SVR & 172 & 0.49 & 0.1332 & 0.1054 & 0.86 & 176 & SVR & 176 &
0.52 & 0.1305 & 0.1025
& 0.83 \\
  & PCFG & 167 & SVR & 167 & 0.27 & 0.1647 & 0.1253 & 0.94 & 175 & SVR & 175 & 0.36 & 0.1614 & 0.1213 & 0.91
\\
  & CJ & 179 & SVR & 179 & 0.27 & 0.103 & 0.0701 & 0.88 & 174 & SVR-FS & 11 & 0.22 & 0.1092 & 0.0682 & 0.86
\\
  \hline
  \multirow{3}{*}{ +C$F_1$ }  & CCL & 173 & SVR & 173 & 0.89 & 0.071 & 0.0517 & 0.42 & 177 & SVR & 177 &
0.89 & 0.0687 & 0.0499
& 0.4 \\
  & PCFG & 168 & SVR & 168 & 0.8 & 0.1027 & 0.0783 & 0.59 & 176 & TREE & 176 & 0.81 & 0.0991 & 0.0754 & 0.56
\\
  & CJ & 180 & SVR & 180 & 0.32 & 0.1009 & 0.0694 & 0.87 & 175 & SVR-PLS & 55 & 0.34 & 0.1036 & 0.0666 &
0.84 \\
  \hline
\end{tabular}
\end{center}
\caption{ID performance on WSJ23 trained on WSJ24 or WSJ0-1-22-24. Using more features or more training data improves. 
The model with the minimum RMSE is selected. \#dimI is the initial number of dimensions of the feature set.}
\label{WSJ23TestResults}
}
\end{table*}

The baseline performances of the parsers over all sentences in the test sets are given in
\Cref{BaselineParserPerformance} where we trained the parsers over WSJ02-21.\footnote{The number of
instances are the same as in~\cite{DBLP:journals/csl/BacchianiRRS06} and
in~\cite{DBLP:conf/emnlp/KummerfeldHCK12} for
WSJ23. The number of sentences reported in \cite{SoricutAPP2008} are lower.}
We also investigate the effect of true casing versus lower casing text when predicting the parser
performance.
CCL lowercases input text and outputs lowercased trees; hence its performance is independent of
casing.\footnote{The output CCL tree is composed of text without labels and to be able to use the
EVALB
bracketing scores, we label each node with 'NP' and enclose them with brackets. We could use any tag
instead
of NP since we are not calculating tag accuracy.}  We observe that CCL's performance slightly
increases on BTest whereas
supervised parsers perform worse.

All of the parsers are trained on WSJ02-21. We use WSJ24 or WSJ0,1,22,24 as training data for the
learning models and
Europarl v7 and News Commentary datasets~\cite{WMT2012} together with the sentences from WSJ0-22,24
training data for training the LM.
We perform prediction experiments on both in-domain (ID) and out-of-domain (OOD) test sets where the
ID test set is
WSJ23 and the OOD test set is BTest. We use the bracketing $F_1$ measure as the target to predict.
Textual features (Text) is the baseline learning setting we use. We then gradually add link
structure based
(+Link), tree structure based (+TreeF), and comparative $F_1$ score based (+C$F_1$) features
for the prediction system to obtain different learning settings.

\subsection{ID Results}

\Cref{WSJ23TestResults} present ID PPP results with WSJ23 as the test set. The first main
column lists the learning setting and the parser.
The second and third main columns present the results we obtain using WSJ24 or WSJ0-1-22-24 as the
training set for the predictor. The model used is either RR, SVR, or TREE and it can be after FS or PLS and we select 
the model with the minimum RMSE for each row of results.
\#dimI is the initial number of dimensions of the feature set being used. Difference between the \#dimI numbers for 
the +TreeF and +Link feature sets correspond to the number of bracketing structural features added, $5$ of which are 
bracketing tree statistical features. +C$F_1$ adds only $1$ feature to the overall feature set.

\begin{table*}[t]
{
\begin{center}
\hspace*{0.5cm}
\begin{tabular}{@{\hspace{0.0cm}}l@{\hspace{0.2cm}}l|c@{\hspace{0.2cm}}r@{\hspace{
0.2cm}}l@{\hspace{0.2cm}}l@{\hspace{0.2cm}}l@{\hspace{0.2cm}}l@{\hspace{0.0cm}}}
Setting & Parser & Model & \#dim & $r$ & RMSE & MAE & RAE \\
\hline
   \multirow{3}{*}{ Text }  & CCL & SVR & 132 & 0.48 & 0.1341 & 0.1068 & 0.87 \\
  & PCFG & SVR & 132 & 0.32 & 0.1623 & 0.1234 & 0.92 \\
  & CJ & SVR-FS & 2 & 0.2 & 0.1059 & 0.0703 & 0.88 \\
  \hline
  \multirow{3}{*}{ +Link }  & CCL & SVR & 158 & 0.49 & 0.1335 & 0.1062 & 0.86 \\
  & PCFG & SVR & 158 & 0.33 & 0.1616 & 0.1231 & 0.92 \\
  & CJ & SVR & 158 & 0.26 & 0.1033 & 0.0706 & 0.89 \\
  \hline
  \multirow{3}{*}{ +TreeF }  & CCL & SVR & 172 & 0.5 & 0.1325 & 0.105 & 0.85 \\
  & PCFG & SVR & 167 & 0.32 & 0.1625 & 0.1234 & 0.92 \\
  & CJ & SVR & 179 & 0.26 & 0.1034 & 0.0706 & 0.89 \\
  \hline
  \multirow{3}{*}{ +C$F_1$ }  & CCL & RR-FS & 3 & 0.89 & 0.0699 & 0.0511 & 0.41 \\
  & PCFG & SVR & 168 & 0.8 & 0.1029 & 0.0784 & 0.59 \\
  & CJ & SVR & 180 & 0.32 & 0.1011 & 0.0692 & 0.87 \\
  \hline
\end{tabular}
\end{center}
}
\caption{Uncased results on WSJ23 trained on WSJ24.}
\label{WSJ23LCTestResults}
\end{table*}

In \Cref{WSJ23TestResults}, the performance improves as we add more features (towards the bottom) or
use more training
data (right) as expected. SVR performs the best with a smaller training set and PLS and FS can
sometimes improve the
performance with a larger training set.
Previous work~\cite{SoricutAPP2008} obtains $0.42$ for $r$ and $0.098$ for RMSE when
predicting the performance of CJ trained on WSJ02-21 over test set WSJ23.
We obtain close $r$ and similar RMSE values with setting +C$F_1$ however, we do not use any parser
or label
dependent information and our C$F_1$ calculation does not use a top performing reference parser
whose performance is
close to CJ's.
Ravi et al.~\cite{SoricutAPP2008} also do not present separate results with the feature sets
they use. The top $r$
they obtain with their text based features is $0.19$, which is about $50\%$ lower than our results
in setting Text.
MAE treats errors equally whereas RMSE is giving more weight to larger errors and can become
dominated by the largest
error. Therefore, MAE and RAE are better metrics to evaluate the performance.
A high RAE indicates that PPP is hard and currently, we can only reduce the error with respect to
knowing and predicting the mean by about $17\%$ for CCL, $14\%$ for CJ, and by $9\%$ for PCFG in the
+TreeF setting.
CCL parsing output is the easiest to predict as we see from the RAE results.
As more additional information external to the given training text is used to derive the parsing
output, the harder it
becomes to predict the performance without supervised label information as we observe for PCFG and
CJ parser outputs.
We are able to predict the performance of CCL better with +Link.
The learning setting +C$F_1$ significantly improves the performance of all prediction results
achieving $r=0.89$,
$0.05$ MAE, and $0.4$ RAE for CCL.
The MAE we achieve in setting +Link is $0.0678$ and it is about $7.4\%$ of the $0.9141$
overall $F_1$ score for CJ on WSJ23.
This error percentage is $18.6\%$ and $17.5\%$ for CCL and PCFG respectively in setting +TreeF.

Uncased in-domain prediction results are given in \Cref{WSJ23LCTestResults} where we train the
parsers using uncased
WSJ02-21, parse uncased WSJ24 and WSJ23, and lowercase the training text before deriving features.
The results show
that uncased results are not necessarily easier to predict.

\begin{table*}[t]
\begin{center}
\small
\begin{tabular}{@{\hspace{0.0cm}}l@{\hspace{0.2cm}}l|c@{\hspace{0.2cm}}@{\hspace{0.2cm
}}c@{\hspace{0.2cm}}r@{\hspace{0.2cm}}l@{\hspace{0.2cm}}l@{\hspace{0.2cm}}l@{\hspace{0.2cm}}l|c@{
\hspace{0.2cm}}c@{\hspace{0.2cm}}r@{\hspace{0.2cm}}l@{\hspace{0.2cm}}l@{\hspace{0.2cm}}l@{\hspace{0.2cm}}l@{\hspace{
0.0cm}}}
& & \multicolumn{7}{c|}{Train: WSJ24} & \multicolumn{7}{c}{Train: WSJ0-1-22-24} \\
Setting & Parser & \#dimI & Model & \#dim & $r$ &
RMSE & MAE & RAE & \#dimI & Model & \#dim & $r$ & RMSE & MAE & RAE \\
\hline
 \multirow{3}{*}{ Text }  & CCL & 132 & SVR & 132 & 0.44 & 0.144 & 0.116 & 0.91 & 132 & SVR & 132 & 0.43
& 0.1452 & 0.1156 &
0.91 \\
  & PCFG & 132 & SVR & 132 & 0.26 & 0.1828 & 0.1387 & 0.93 & 132 & SVR & 132 & 0.23 & 0.1865 & 0.1403 & 0.95
\\
  & CJ & 132 & SVR & 132 & 0.27 & 0.1667 & 0.1037 & 0.87 & 132 & SVR & 132 & 0.26 & 0.1719 & 0.1039 & 0.87
\\
  \hline
  \multirow{3}{*}{ +Link }  & CCL & 158 & SVR & 158 & 0.45 & 0.1433 & 0.1155 & 0.91 & 158 & SVR & 158 &
0.45 & 0.144 & 0.1151 &
0.91 \\
  & PCFG & 158 & SVR & 158 & 0.28 & 0.1814 & 0.1383 & 0.93 & 158 & SVR & 158 & 0.25 & 0.1841 & 0.139 & 0.94
\\
  & CJ & 158 & SVR & 158 & 0.25 & 0.1669 & 0.1039 & 0.87 & 158 & SVR & 158 & 0.26 & 0.1707 & 0.1036 & 0.87
\\
  \hline
  \multirow{3}{*}{ +TreeF }  & CCL & 172 & SVR & 172 & 0.45 & 0.1437 & 0.1153 & 0.91 & 176 & SVR & 176 &
0.46 & 0.1428 & 0.1137
& 0.9 \\
  & PCFG & 167 & SVR & 167 & 0.23 & 0.1837 & 0.1397 & 0.94 & 175 & SVR & 175 & 0.25 & 0.1848 & 0.1394 & 0.94
\\
  & CJ & 179 & SVR & 179 & 0.27 & 0.1661 & 0.1033 & 0.87 & 174 & SVR & 174 & 0.26 & 0.1698 & 0.1029 & 0.86
\\
  \hline
  \multirow{3}{*}{ +C$F_1$ }  & CCL & 173 & RR & 173 & 0.84 & 0.0918 & 0.067 & 0.53 & 177 & RR & 177 &
0.84 & 0.0909 & 0.0661 &
0.52 \\
  & PCFG & 168 & SVR & 168 & 0.76 & 0.1213 & 0.0908 & 0.61 & 176 & SVR-PLS & 39 & 0.77 & 0.1205 & 0.089 &
0.6 \\
  & CJ & 180 & SVR & 180 & 0.34 & 0.1616 & 0.1013 & 0.85 & 175 & RR-PLS & 30 & 0.38 & 0.1542 & 0.1 & 0.84 \\
  \hline
\end{tabular}
\end{center}
\caption{OOD prediction performance on BTest trained on WSJ24 or WSJ0-1-22-24.}
\label{BTestTestResults}
\end{table*}

\subsection{OOD Results}

We use the BTest corpus for PPP in an OOD scenario. OOD parsing decreases the performance of
supervised parsers as is
given in \Cref{BaselineParserPerformance}. However, this is not the case for unsupervised parsers
such as CCL since
they use limited domain dependent information and in fact CCL's performance is slightly increased.
OOD PPP results are
given in \Cref{BTestTestResults}. We observe that prediction performance is lower when we compare
OOD results with ID
results.
In OOD learning, the addition of the TreeF feature set improves the performance more when compared
with the improvement in ID.
Previous work~\cite{SoricutAPP2008} obtains $0.129$ for RMSE when predicting the performance of CJ
trained on WSJ02-21
over BTest. We obtain slightly worse RMSE values with setting +C$F_1$. We also note that the number
of sentences
reported in \cite{SoricutAPP2008} for datasets WSJ23, WSJ24, and BTest are lower than the official
datasets released as
part of Penn Treebank~\cite{PennTB1993}.
Our prediction of CCL or PCFG performance achieve better RAE levels of $0.5$ and $0.6$ respectively in OOD than 
sentence MTPP.

\begin{table*}[t]
\centering
{\small
\begin{tabular}{@{\hspace{0.0cm}}cc|c@{\hspace{0.1cm}}r|c@{\hspace{0.1cm}}r|c@{\hspace{0.1cm}}r|c@{\hspace{0.1cm}}r@{
\hspace{0.0cm}}}
& & \multicolumn{2}{c}{Text} & \multicolumn{2}{c}{+Link} & \multicolumn{2}{c}{+TreeF} & \multicolumn{2}{c}{+C$F_1$} \\
 \hline
 \multirow{6}{*}{\begin{sideways}wsj24\end{sideways}} & \multirow{2}{*}{CCL}   & 
lensS & -0.403 & b$5$gram ppl & 0.009 & b$5$gram ppl & 0.009 & C$F_1$ & 0.885\\
 &   & b$2$gram logp & 0.429 & $5$gram ppl & -0.001 & $5$gram ppl & -0.001 & 2gram BLEU bnd & 0.052\\
\cline{2-10}
 & \multirow{2}{*}{PCFG}   & 
b$5$gram ppl & -0.017 & b$5$gram ppl & -0.017 & b$5$gram ppl & -0.017 & C$F_1$ & 0.753\\
 &   & $5$gram ppl & -0.023 & b$2$gram logp & 0.172 & $5$gram ppl & -0.023 & 2gram $F_1$ bnd & 0.054\\
\cline{2-10}
& \multirow{2}{*}{CJ}   & 
b$5$gram ppl & 0.013 & b$5$gram ppl & 0.013 & $5$gram ppl & 0.003 & 3gram $F_1$ bnd & 0.094\\
 &   &  &  & $5$gram ppl & 0.003 & $3$gram f & -0.184 &  & \\
\hline
\multirow{6}{*}{\begin{sideways}wsj0-1-22-24\end{sideways}} & 
\multirow{2}{*}{CCL}   & max logp3 & 0.058 & max logp3 & -0.343 & max logp3 & -0.343 & 
max logp3 & -0.343\\
 &   & max logp2 & -0.320 & max logp2 & -0.330 & max logp2 & -0.330 & max logp2 & -0.330\\
\cline{2-10}
& \multirow{2}{*}{PCFG}   & 
max logpj3 bpw & 0.023 & max logpj3 bpw & -0.102 & max logpj3 bpw & -0.102 & max logp2 & -0.095\\
 &   & max logpj2 bpw & -0.066 & max logpj2 bpw & -0.099 & max logpj2 bpw & -0.099 & max logp3 & -0.088\\
\cline{2-10}
 & \multirow{2}{*}{CJ}   & 
max logp3 & 0.069 & max logpj3 & 0.173 & max logpj3 & 0.173 & max logpj3 & 0.173\\
 &   & max logp2 & -0.104 & max logpj2 & 0.173 & max logpj2 & 0.173 & max logpj2 & 0.173\\
\cline{2-6}
\hline
\end{tabular}
}\caption{PPP top $2$ features for SVR when predicting CCL, PCFG, or CJ with training set either wsj24 or wsj0-1-22-24.}
\label{TopFeatures}
\end{table*}

\subsection{Feature Selection Results}
\label{FeatureSelectionResults}

The ranking of the features we obtain after FS are given in \Cref{TopFeatures} where up to top $5$ features
for each learning setting is presented when predicting PCFG. As expected, the C$F_1$ ranks high in setting +C$F_1$. 
Interestingly, tree features are not among the top features selected. 
We observe that translation and LM based features are ranked high.
Features using link structures are also abundant among the top features.
In contrast to previous work~\cite{SoricutAPP2008}, number of OOV words are not selected among the
top features. Abbreviations used are as follows:
A number corresponds to the order of the $n$-grams used or the LM order, \textit{len} is the length of a sentence 
in the number of words, ppl is perplaxity, and GM is the geometric mean between the precision and recall.
\textit{b$1$gram logp} is the backward $1$-gram log probability.
3gram w$F_1$ is weighted $F_1$ score over $3$-gram features weighted according to the sum of the likelihood of 
observing them among $3$-grams. link w$F_1$ is over links from CCL. \textit{wrec} is weighted recall.
\textit{max logp$k$} is the Bayes IBM model 1~\cite{Brown93} translation log probability for top $k$ translations (we 
omit $k$ when $k=1$). \textit{bpw} correspond to the LM bits per word.

\section{Contributions}

MTPPS-PPP works without training a parser, without parsing with it, without any parser dependent 
information, and without looking at the parsing output.
We have obtained SoA PPP results and obtained expected lower bound on the prediction performance and the number of 
instances needed for prediction given a RAE level, which can be useful for building SoA predictors for a given 
prediction task. 
MTPPS-PPP requires minutes for PPP and achieves $0.85$ RAE.
Our prediction results allow better setting of expectations for each task and domain.
We also provide ranking of the features used for different domains and tasks.
Ability to predict outcomes enables preparation and savings in computational effort, which may determine whether a 
business can survive or not in industrial settings. 
Our results show that we only need 11 labeled instances for PPP to reach SoA prediction performance.

\section*{Acknowledgments}

This work is supported in part by SFI (13/TIDA/I2740) for the project ``Monolingual and Bilingual Text Quality Judgments 
with Translation Performance Prediction'' ({\scriptsize \texttt{computing.dcu.ie/\textasciitilde 
ebicici/Projects/TIDA\_RTM.html}}) and in part by SFI (12/CE/I2267) as part of the
ADAPT CNGL Centre for Global Intelligent Content ({\scriptsize \texttt{adaptcentre.ie}}) at Dublin City University.

\bibliography{CNGL}

\end{document}